\documentclass[runningheads]{llncs}
\usepackage{graphicx}

\usepackage{multirow}
\usepackage{amssymb}
\usepackage{multicol}
\usepackage{amsmath}

\usepackage{xspace} 
\usepackage{tabularx}
\usepackage{subcaption}

\usepackage{array,booktabs}
\newcommand {\otoprule}{\midrule [\heavyrulewidth]}
\newcolumntype {+}{ >{\global\let\currentrowstyle\relax}}
\newcolumntype {^}{ >{\currentrowstyle }}
  \newcommand {\rowstyle}[1]{\gdef\currentrowstyle{#1} %
  #1\ignorespaces
  }
\newcommand{\tabhead}{\rowstyle{\bfseries}}
\usepackage[para,online,flushleft]{threeparttable}
\usepackage[T1]{fontenc}
\usepackage{pifont}
\usepackage{tablefootnote}

\newcommand{\cbis}{CBIS\xspace}

\begin{document}
\title{Prototype-based Interpretable Breast Cancer Prediction Models: Analysis and Challenges}
\titlerunning{Prototype-based Interpretable Breast Cancer Prediction Models}
\author{Shreyasi Pathak\thanks{Corresponding author}\inst{1}\orcidID{0000-0002-6984-8208} \and
Jörg Schlötterer\inst{2,4}\orcidID{0000-0002-3678-0390} \and
Jeroen Veltman\inst{3}\orcidID{0000-0002-6824-3987} \and
Jeroen Geerdink\inst{3}\orcidID{0000-0001-6718-6653} \and
Maurice van Keulen\inst{1}\orcidID{0000-0003-2436-1372} \and
Christin Seifert\inst{2}\orcidID{0000-0002-6776-3868}}
\authorrunning{S. Pathak et al.}
\institute{University of Twente, The Netherlands \and
Philipps-University Marburg, Germany \and
Hospital Group Twente (ZGT), The Netherlands \and
University of Mannheim, Germany\\
\email{\{s.pathak,m.vankeulen\}@utwente.nl}
\email{\{christin.seifert,joerg.schloetterer\}@uni-marburg.de}
\email{\{j.veltman,j.geerdink\}@zgt.nl}}

\maketitle              %
\begin{abstract}

Deep learning models have achieved high performance in medical applications, however, their adoption in clinical practice is hindered due to their black-box nature. Using explainable AI (XAI) in high-stake medical decisions could increase their usability in clinical settings. Self-explainable models, like prototype-based models, can be especially beneficial as they are interpretable by design. However, if the learnt prototypes are of low quality then the prototype-based models are as good as black-box. Having high quality prototypes is a pre-requisite for a truly interpretable model. In this work, we propose a prototype evaluation framework for Coherence (PEF-Coh) for quantitatively evaluating the quality of the prototypes based on domain knowledge. We show the use of PEF-Coh in the context of breast cancer prediction using mammography. Existing works on prototype-based models on breast cancer prediction using mammography have focused on improving the classification performance of prototype-based models compared to black-box models and have evaluated prototype quality through anecdotal evidence. We are the first to go beyond anecdotal evidence and evaluate the quality of the mammography prototypes systematically using our PEF-Coh. Specifically, we apply three state-of-the-art prototype-based models, ProtoPNet, BRAIxProtoPNet++ and PIP-Net on mammography images for breast cancer prediction and evaluate these models w.r.t. i) classification performance, and ii) quality of the prototypes, on three public datasets. Our results show that prototype-based models are competitive with black-box models in terms of classification performance, and achieve a higher score in detecting ROIs. However, the quality of the prototypes are not yet sufficient and can be improved in aspects of relevance, purity and learning a variety of prototypes. We call the XAI community to systematically evaluate the quality of the prototypes to check their true usability in high stake decisions and improve such models further.

\keywords{Explainable AI  \and Prototype-based models \and Breast cancer prediction \and Mammography.}
\end{abstract}
   
\section{Introduction}
\label{sec:intro}

Deep learning has achieved high performance on various medical applications, e.g. breast cancer prediction~\cite{shen2021interpretable,shen2019deep}. However, its adoption to clinical practice is hindered due to its black-box nature. Explainable AI (XAI) aims to address this gap by explaining the reasoning of black-box models in a post-hoc manner (post-hoc explainable methods) or developing models which are interpretable by design (self-explainable methods). 

Post-hoc explainable methods explain a trained deep neural network with a separate explainer model, e.g. Grad-CAM can highlight the important image regions to explain the prediction from a neural network~\cite{oh2020deep}. However, these explanations are not always reliable or trustworthy~\cite{rudin2019stop}, as they come from a second explainer model.
Self-explainable models were developed to address this issue -- one of the most prominent ones being prototype-based models, e.g. ProtoPNet~\cite{chen2019looks}.

Breast cancer prediction using mammography is a challenging medical task as the regions-of-interest (ROIs) are small compared to the whole mammogram and presence of dense breast tissue may make it difficult to read the mammogram. Depending on the finding it can be difficult for a radiologist to decide whether to perform a biopsy, wait for follow up or diagnose the lesion as certainly benign.
Learning relevant malignant and benign prototypes from the training data and showing the reasoning behind a model's prediction might assist radiologists in taking data based decision. Existing works~\cite{wang2022knowledge,wang2023interpretable} have extended prototype-based model, ProtoPNet~\cite{chen2019looks} on breast cancer prediction using whole mammography images.~\cite{barnett2021case} extended ProtoPNet for breast cancer prediction on ROIs from mammography images. However, most works show anecdotal evidence for the explanations. The quality of the learnt prototypes from mammography images have not been extensively evaluated yet.

A recent survey~\cite{nauta2023anecdotal} found that most XAI methods are usually evaluated based on anecdotal evidence by selecting a few good examples. However, anectodal evidence is not sufficient to judge the quality of the explanations. Nauta et al.~\cite{nauta2023anecdotal} proposed a framework of twelve properties for evaluating the quality of explanations. One of the properties, \textit{Coherence}, evaluates whether the explanations are correct with respect to domain knowledge. In the medical domain, it is  highly important to evaluate the explanation quality in terms of Coherence, before the model can be adopted in a clinical setting.

In this work, we go beyond anecdotal evidence and propose a \textit{Prototype Evaluation Framework} (PEF), with special emphasis on \textit{Coherence} (PEF-Coh) to quantitatively evaluate the quality of prototypes and to provide a broader evaluation of prototype-based models based on the domain knowledge. Specifically, we apply state-of-the-art prototype based models on breast cancer prediction using mammography images and evaluate the quality of the prototypes both quantitatively and qualitatively. Our contributions are as follows:
\begin{enumerate}
    \item We propose PEF-Coh, a Prototype Evaluation Framework for Coherence, for evaluating the quality of prototypes in prototype-based models. Our framework can be used as a comprehensible and generalizable evaluation framework for prototype-based models of medical applications.  
    \item We reproduced a state-of-the-art breast cancer prediction model, BRAIxProtoPNet++, which had no public source code. We release the source code of the evaluation framework, the implementations of all black-box\footnote{\label{fn:repo-blackbox}\url{https://github.com/ShreyasiPathak/multiinstance-learning-mammography}} and interpretable models\footnote{\label{fn:repo-prototype}\url{https://github.com/ShreyasiPathak/prototype-based-breast-cancer-prediction}} for further research.
    \item We systematically compare prototype-based models and blackbox models on three standard benchmark datasets for breast cancer detection, and show that interpretable models achieve comparative performance. Our analysis of prototype quality identifies differences between prototype-based models and points towards promising research directions to further improve model's coherence.
\end{enumerate}

\section{Related Work}
\label{sec:relwork}

\textbf{Prototype-based models.} Prototype-based models were first introduced with ProtoPNet~\cite{chen2019looks}, an intrinsically interpretable model with similar performance to black-box models. ProtoTree~\cite{nauta2021neural} reduces the number of prototypes in ProtoPNet and uses a decision tree structure instead of a linear combination of prototypes as decision layer. ProtoPShare~\cite{rymarczyk2021protopshare} and ProtoPool~\cite{rymarczyk2022interpretable} also optimize towards reducing the number of prototypes used in classification. TesNet~\cite{wang2021interpretable} disentangles the latent space of ProtoPNet by learning prototypes on a Grassman manifold. PIP-Net~\cite{nauta2023pip} addresses the semantic gap between latent space and pixel space for the prototypes, adds sparsity to the classification layer and can handle out-of-distribution data. XProtoNet~\cite{kim2021xprotonet} is a prototype-based model developed for chest radiography and achieved state-of-the-art classification performance on a chest X-ray dataset.

\textbf{Interpretability in Breast Cancer Prediction.}~\cite{wu2018expert} developed a expert-in-a-loop interpretation method for interpreting and labelling the internal representation of the convolutional neural network for mammography classification.  BRAIxProtoPNet++~\cite{wang2022knowledge} extended ProtoPNet for breast cancer prediction achieving a higher performance compared to ProtoPNet and outperforming black-box models in classification performance. InterNRL~\cite{wang2023interpretable} extended on BRAIxProtoPNet++ using a student-teacher reciprocal learning approach. IAIA-BL~\cite{barnett2021case} extended ProtoPNet with some supervised training on fine-grained annotations and applied on ROIs instead of whole mammogram images. 

\textbf{Evaluating XAI.} Most XAI methods are usually evaluated using anecdotal evidence~\cite{nauta2023anecdotal}. However, the field is moving towards standardizing evaluation of explanations similar to the standardized approach for evaluating classification performance, e.g. Co-12 property~\cite{nauta2023anecdotal,nauta2023co}. Nauta et al.~\cite{nauta2023co} reviewed how Co-12 properties are evaluated in existing works with prototype-based models and found one of the Co-12 properties, Coherence is often evaluated with anecdotal evidence and sometimes with objective measures like localization~\cite{xu2023sanity} and purity~\cite{nauta2023pip}. To the best of our knowledge, this is the first work to introduce a Prototype Evaluation Framework for Coherence (PEF-Coh) and systematically evaluate the quality of prototypes in the context of breast cancer prediction. We use ProtoPNet, BRAIxProtoPNet++ and PIP-Net on breast cancer prediction using mammography images and evaluate on three public datasets - CBIS-DDSM~\cite{cbisddsm}, VinDr~\cite{Nguyenvindr} and CMMD~\cite{cmmd} for classification performance. We use PEF-Coh to evaluate the quality of the learnt prototypes of the three models using the fine-grained annotations of abnormality types available in CBIS-DDSM~\cite{cbisddsm} dataset.

\section{Preliminaries}
In this section, we provide a summary of the state-of-the-art models used in this paper to make this paper self-contained.

\subsection*{Problem Statement}
We pose breast cancer prediction using mammography as a classification problem. We define it as a binary classification problem for the datasets, CBIS-DDSM~\cite{cbisddsm} and CMMD~\cite{cmmd} with 2 classes, benign and malignant and a multiclass classification problem for the dataset, VinDr~\cite{Nguyenvindr}, containing the 5 BI-RADS~\cite{sickles2013acr} categories (cf. Sec. \ref{sec:expsetupdataset}). Suppose, $\{(x_1, y_1), $\ldots$, (x_n, y_n), $\ldots$, (x_N, y_N)\} \in \mathcal{X} \times \mathcal{Y}$ is a set of N training images with each image having a class label $y_n \in \{0,1\}$ ($0$: benign, $1$: malignant) for binary classification and a class label $y_n \in \{0,1,2,3,4\}$ ($0$: BI-RADS 1, $1$: BI-RADS 2, $2$: BI-RADS 3, $3$: BI-RADS 4, $4$: BI-RADS 5) for multiclass classification. Our objective is to learn the prediction function $f$ in $\hat{y}_n = f(x_n)$ where $\hat{y}_n$ is the predicted label.

\subsection*{Black-box Models}
We used pretrained \textit{ConvNext}~\cite{liu2022convnet} and \textit{EfficientNet}~\cite{tan2019efficientnet} models and changed the classification layer to 2 classes and 5 classes with a softmax activation function. 
\textit{GMIC}~\cite{shen2021interpretable} is a state-of-the-art breast cancer prediction model containing a global module with feature extractor for learning the features from a whole image, a local module with feature extractor for learning the features from the patches and a fusion module to concatenate the global and local features. The global module generates a saliency map, which is used to retrieve the top-k ROI candidates. Local module takes these ROI candidates (patches) as input and aggregates the features using a multi-instance learning method. The global, local and fusion features are separately used for prediction and passed to the loss function for training GMIC. During inference, only fusion features are used for prediction. We used their model for binary classification with 1 neuron in the classification layer with a sigmoid activation function. For multiclass classification, we used their setup of multilabel classification and replaced sigmoid with a softmax activation function. 

\subsection*{Prototye-based Models}
\textit{ProtoPNet}~\cite{chen2019looks} consists of a convolutional feature extractor backbone, a layer with prototype vectors and a classification layer. The feature extractor backbone generates a feature map of $H \times W \times D$ where H, W, and D are height, width and depth of the feature map. A prototype vector of size $1 \times 1 \times D$ is passed over $H \times W$ spatial locations to find the patch in the feature map that is most similar to the prototype vector. The prototype layer contains equal number of prototype vectors for each class. In the classification layer, the prototype vectors are connected to their own class with a weight of 1 and to other classes with a weight of -0.5. After the most similar patch for each prototype is saved, the patch is visualized in the pixel space using upsampling. 

\textit{BRAIxProtoPNet++}~\cite{wang2022knowledge} extends on ProtoPNet to improve the classification performance by including a global network and distilling the knowledge from the global network to the ProtoPNet. It also increases the prototype diversity by taking all prototypes from different training images. 

\textit{PIP-Net}~\cite{nauta2023pip} does not use a separate prototypical layer like ProtoPNet and BRAIxProtoPNet++, but uses regularization to make the features in the last layer of the convolutional backbone interpretable. For a feature map of dimension $H \times W \times D$, the model can have a maximum of $D$ prototypes. Each patch of dimension $1 \times 1 \times D$ in the feature map is softmaxed over $D$ such that one patch tries to match to maximally one prototype. The maximum similarity score is taken from each of the $D$ channels, which is then used in the classification layer. Further, the classification layer is optimized for sparsity to keep only the positive connections towards a decision.     

\begin{table*}[t!]
\setlength{\tabcolsep}{5pt}
\centering
\caption{Overview of our prototype evaluation framework (PEF) along with the associated measure. We also show the relation of PEF properties to the Co-12 framework~\cite{nauta2023anecdotal}. Global prototypes (GP) is the number of prototypes with non-zero weights in the classification layer, local prototypes (LP) is the number of prototypes active per instance, relevant prototypes (RP) contain ROIs. Categories refer to fine-grained part annotations from a lexicon (abnormality type and BI-RADS descriptors in our use case). Total number of categories (TC) are all possible categories from the lexicon, unique categories (UC) is the size of the set of categories found by the model. Level refers to whether the measure is calculated at the local (L), i.e., at the image-level or global 
(G) model level.}
\label{tab:approach:evaluation-framework}
\begin{small}
\begin{tabularx}{\textwidth}{+l^l^X^c^l} 
\toprule\tabhead
Co-12  & PEF & Question & Level & Measure  \\ 
\textbf{property}~\cite{nauta2023anecdotal} & \textbf{property} &  &  &   \\ \otoprule
Compactness & Compactness & What is the explanation size? & G & $GP$\\
   & & & L & $LP$ \\
Coherence & Relevance & How many of the learnt prototypes are relevant for diagnosis? & G & $|RP|/GP$ \\
Coherence & Specialization & Are the prototypes meaningful enough to be assigned names of abnormality categories? & G & Purity$_{RP}/|RP|$ \\ 
Coherence & Uniqueness & Does the model activate more than one prototype for one abnormality category? & G & $UC_{RP}/|RP|$ \\
Coherence & Coverage & How many abnormality categories was the model able to learn? & G & $UC_{RP}/TC$ \\
Coherence & Localization & To what extent is the model able to find correct ROIs? & L & IoU, DSC   \\
Coherence & Class-specific & Is the model associating the prototypes with the correct class? & G & 
$\frac{Align(W_{RP},CD_{C_{RP}})}{|RP|}$\\
\bottomrule
\end{tabularx}
\end{small}
\end{table*}

\section{Prototype Evaluation Framework}
\label{sec:approach}
We developed a \textit{Prototype Evaluation Framework} (PEF) with special emphasis on \textit{Coherence}~\cite{nauta2023anecdotal} (PEF-Coh) to i) quantitatively evaluate the quality of the learnt prototypes and ii) provide a broader assessment of the prototype-based models, based on domain knowledge. We use our PEF to assess the status quo of the prototype-based models on breast cancer prediction using mammography. Our PEF contains 7 properties, including 1 property measuring Compactness and 6 properties measuring Coherence, summarized in Table~\ref{tab:approach:evaluation-framework}: 

\begin{enumerate}
\item \textbf{Compactness} measures the size of the local and global explanation of the model. The lower the explanation size, the easier it is to comprehend for users. Note that this is the only property in our PEF that belongs to the Co-12~\cite{nauta2023anecdotal} property Compactness and not to Coherence. We decided to include it in our PEF as the measurement of the Coherence properties defined later is based on the explanation size, so Coherence related measures are effected by the Compactness measure. 
We calculate this as follows:\\
\textit{Global prototypes (GP):} Total number of prototypes with non-zero weights to any class in the classification layer.\\
\textit{Local prototypes (LP):} Average number of prototypes that are active per instance, i.e., whose contribution to the classification decision (prototype presence score $p$ times weight $w$ to the classification layer) is non-zero.

\item \textbf{Relevance} measures the number of prototypes associated with the ROI\footnote{While ROI is most commonly used in the medical domain, we use it as generic term to refer to any annotated part of an image, including object part annotations.} among the global prototypes. The larger the number of relevant prototypes, the better the capability of the model in finding the relevant regions. We measure this property by calculating the ratio between the number of prototypes activated on at least one ROI (relevant prototypes (RP)) and the number of global prototypes (cf. Eq.~\ref{eq:1}): 
\begin{equation} 
Rel. = \frac{|RP|}{GP}
\label{eq:1}
\end{equation}
Specifically, to determine a relevant prototype, $RP$, we take the top-k training instances with highest prototype presence scores for each prototype (i.e., images, where the prototype is most activated). If the corresponding patch for the prototype in any of these top-k instances overlaps with the center of a ROI in this image, we consider it relevant.

\item \textbf{Specialization} measures the purity of prototypes by matching the extent of its alignment to lexical categories (cf. Eq.~\ref{eq:3}). A truly interpretable prototype represents a single semantic concept that is comprehensible to humans, i.e., can be named or given a meaning by users. For example in our use case, a prototype is not fully interpretable if it gets activated both on mass or calcification abnormality type, as we don't know whether to call this prototype a mass prototype or a calcification prototype. The higher the purity, the more interpretable the prototypes.
\begin{equation}
Spec.=\frac{\sum_{p \in RP}Purity_{p}}{|RP|}
\label{eq:3}
\end{equation}
where $RP$ is the set of relevant prototypes as defined above and $Purity_{p}$ is the share of the majority category in top-k patches (relevant prototypes overlap with an ROI center and each ROI is labelled with a category). If all top-k patches belong to the same category, the prototype is pure.
For a detailed evaluation by category, we assign the majority category and report purity scores per category. 
If the category lexicon constitutes a hierarchy like in our use case, we can determine purity also on more fine-grained category levels. 

\item \textbf{Uniqueness} measures the number of unique categories of prototypes among all relevant prototypes, formally defined in Eq.~\ref{eq:4}. 
\begin{equation}
Uniq. = \frac{UC_{RP}}{|RP|}
\label{eq:4}
\end{equation}
where $UC_{RP}$ stands for unique categories (UC) of all relevant prototypes. \\
Each prototype has a single (majority) category assigned from the purity calculation of the Specialization measure before. Ideally, we want each prototype to represent a different, unique category and not multiple prototypes that represent the same.
The uniqueness measure should always be evaluated in combination with Relevance and Coverage (next measure), as trivially, a single relevant prototype would yield the maximum uniqueness score of 1.

\item \textbf{Coverage} 
measures the fraction of all categories in the dataset that are represented by prototypes, formally defined in Eq.~\ref{eq:5}. 
\begin{equation}
Cov. = \frac{UC_{RP}}{TC}
\label{eq:5}
\end{equation}
where $TC$ is the total number of categories. \\
Ideally, a model has at least one prototype associated with each category, such that it is able to incorporate all variations in its reasoning process.

\item \textbf{Localization} measures the capability of the model in finding the correct ROIs through the activated prototypes, formally defined in Eq.~\ref{eq:6}. 
\begin{equation}
Loc. = \frac{\sum_{t=1}^{T} IoU(ROI_{t}, Patch_{t})}{T}
\label{eq:6}
\end{equation}
where IoU is the intersection over union (IoU) over groundtruth ROI and the patch activated by the prototype, $t$ stands for the $t^{th}$ test instance of the total $T$ test instances. IoU can also be replaced with dice similarity coefficient (DSC). We calculate and report both in our experiments.\\
This property is a local measure, meaning that it is calculated per test instance and averaged over the whole test set. The higher the localization, the better the model is at finding the correct ROIs. 

\item \textbf{Class-specific} measures the extent to which prototypes are associated  with the correct class, formally defined in Eq.~\ref{eq:7}. 
For example, a prototype that has an assigned category (cf. Purity calculation in Specialization measure) that is more often found to belong to a certain class in the dataset, should have a higher weight to that class in the classification layer.
A high class-specific score means the model is good at associating categories with the correct class.
We measure this property by the following:
\begin{equation}
Cl\text{-}spec. = \frac{\sum_{p \in RP}Align(W_{p},CD_{C_{p}})}{|RP|}
\label{eq:7}
\end{equation}
where $Align$ stands for the alignment between the weights of a relevant prototype towards the classes in the classification layer, $W_{RP}$, and the class distribution (CD) of the category (C) assigned to this prototype, $CD_{C_{RP}}$. 
$Align$ is a binary assessment: If a prototype has the highest weight to the majority class of its category, the score is 1 and 0 otherwise. The majority class is obtained from ROI category annotations in the dataset (category class distribution $CD_{C_{RP}}$).
\end{enumerate}

\section{Experimental Setup}
\label{sec:expsetup}

In this section, we describe the datasets, training details, setup of our prototype evaluation framework and the visualization method of the prototypes. 

\subsection{Datasets}
\label{sec:expsetupdataset}
We performed our experiments on three commonly used mammography datasets:\\
\textbf{CBIS-DDSM}~\cite{cbisddsm} is a public dataset containing 3,103 mammograms from 1,566 patients including mediolateral oblique (MLO) and craniocaudal (CC) views. The dataset contains 1,375 malignant and 1,728 benign images. For our experiments, we used the official train (2458 images) and test split (645 images) provided by the dataset.\\
\textbf{CMMD}~\cite{cmmd} is a Chinese public dataset collected between years 2012 and 2016, containing 5,202 mammograms of CC and MLO views from 1,775 patients. It contains 2,632 malignant and 2,570 benign images. We split the dataset in a patient-wise manner into training set (3,200 images, D1-0001 to D2-0247) and test set (2,002 images, D2-0248 to D2-0749) following~\cite{wang2023interpretable} for fair comparison to their results.\\   
\textbf{VinDr-Mammo}~\cite{Nguyenvindr} is a public dataset from two hospitals in Vietnam, containing 5,000 mammography cases (20,000 mammogram images) including CC and MLO views. The images are annotated with BI-RADS category 1 to 5 with 1 being normal mammogram and 5 being most malignant. We used the official training (16,000 images) and test (4,000 images) split of the dataset for our experiments. 

\subsection{Training Details}
We split the training set patient-wise into 90\% train and 10\% validation for all datasets. We performed hyperparameter tuning on CMMD (and when needed also on CBIS-DDSM) dataset and the hyperparameters with the highest validation AUC were selected as the hyperparameter for our experiments on all datasets. The models were trained till a certain epoch (details in the following paragraph) and the model of the epoch with the highest validation AUC was selected for test set inference.

\textit{EfficientNet-B0}~\cite{tan2019efficientnet} was trained with a fixed learning rate (lr) of 0.001 and weight decay (wd) of 1e-5 for 60 epochs. \textit{ConvNext}~\cite{liu2022convnet} was trained with a fixed lr=1e-5 and wd=1e-5 for 50 epochs. \textit{GMIC}~\cite{shen2021interpretable} was trained with a fixed learning rate $\approx$ 6e-5 and weight decay $\approx$ 3e-4 (exact values can be found in our repository\textsuperscript{\ref{fn:repo-blackbox}}) for 50 epochs.
\textit{ProtoPNet}~\cite{chen2019looks} was trained in 3 stages - i) warm optimizer stage where only the add on layer and the prototype vectors are trained with lr=1e-4 for 5 epochs, ii) fine tuning stage where the whole network except the last layer is trained for 55 epochs with lr=1e-5 for the backbone, lr=1e-4 for the add on layers and the prototype vector, iii) the last layer is trained every 11th epoch for 20 iterations with lr=1e-2. \textit{BRAIxProtoPNet++}~\cite{wang2022knowledge} was trained in 3 stages - i) GlobalNet training where the GlobalNet and its classification layer was trained with lr=1e-5 for 3 epochs, ii) ProtoPNet training where the add on layer, the prototype vector and the ProtoPNet classification layer was trained with lr=1e-5 for 3 epochs, and iii) whole model fine-tuning stage, where all the layers were trained with lr=1e-5 for the backbone, lr=1e-4 for the add on layer and the prototype vector and lr=1e-2 for the classification layers for 54 epochs. This schedule of BRAIxProtoPNet++ was followed for CBIS-DDSM dataset, however, for CMMD and VinDr, the (iii) stage used a fixed lr=1e-5 for all layers. 
In \textit{PIP-Net}~\cite{nauta2023pip}, the feature extractor backbone was trained with a fixed lr=0.00001 and the classifier layer was trained with a lr=0.05 in a cosine annealing with warm restart schedule. PIP-Net was pretrained in a self-supervised manner for 10 epochs, followed by full finetuning for 60 epochs following~\cite{nauta2023pip}. 

All images were resized to 1536x768 following~\cite{wang2022knowledge,wang2023interpretable}. Data augmentation of random horizontal flipping, shearing, translation and rotation was applied (details can be found in our repository\textsuperscript{\ref{fn:repo-blackbox},\ref{fn:repo-prototype}}).

We report the average performance and standard deviation over three model runs with random weight initializations with different seeds. We report F1 and AUC of the positive class (malignant) for binary classification task and weighted macro F1 (F1$^{wM}$) and one-vs-one macro AUC (AUC$^M$) for multiclass classification. 

\subsection{Prototype Evaluation Framework Setup}
We calculated the seven properties of our PEF for the three prototype-based models, ProtoPNet, BRAIxProtoPNet++ and PIP-Net on breast cancer prediction task. All analysis were perfomed on CBIS-DDSM, because it is the only dataset with fine-grained annotations of abnormality types -- mass and calcification; mass BI-RADS descriptors - shape and margin; and calcification BI-RADS descriptors - morphology and distribution. We took the top-$k$, where $k=10$ similar patches from the training set for each prototype to calculate the Relevance, Specialization, Uniqueness and Coverage. We set $k=10$ to focus on the most representative instances, but still allow a bit of variation (outliers).

For \textit{Relevance} measure, we checked if any of the top-k patches contain the center of the ROI. If there's atleast 1 match, we mark that prototype as relevant (i.e. prototype associated with ROI). We divide the number of relevant prototypes with the global prototypes of each model to get the relevance score.

For purity calculation in \textit{Specialization} measure, we calculate how many top-k patches contain the center of the ROI in the image, where the patch is from and divide this with $k$ to get the purity score. We calculate the purity score for all abnormality categories that the top-k patches match to and assign the abnormality category with the highest purity score to that prototype. We average the purity score over all relevant prototypes to get the specialized score of the model. We define categories following the hierarchical level of BI-RADS lexicon -- at the top level, abnormality type is the category, i.e. mass or calcification. At a sub-level, mass abnormalities (lesions) can be described with or categorized by shape and margin  descriptors. At the same level, calcification abnormalities can be described with or categorized by a morphology and distribution.

For \textit{Uniqueness}, we define categories by combining the descriptors at all levels. A category for mass is defined with a combination of 
\textit{abnormality type - shape - margin} and for calcification, with a combination of \textit{abnormality type - morphology - distribution}. For example, an abnormality category of a mass can be mass-oval-circumscribed, i.e. a mass with oval shape and circumscribed margin, and an abnormality category for a calcification can be calcification-pleomorphic-segmental, i.e. a calcification with pleomorphic morphology and segmental distribution. Following this rule, we found that the total abnormality categories in the CBIS-DDSM is 132 (70 mass categories, and 62 calcification categories) and we used this value for calculation of \textit{Coverage} property.

For calculating the \textit{Class-specific} measure, we selected the abnormality categories that have instances for both malignant and benign classes in the dataset. 

For \textit{Localization}, we calculated the Intersection over Union (IoU) and Dice Similarity Coefficient (DSC) for the top-1 (IoU$^{1}$, DSC$^{1}$), top-10 (IoU$^{10}$, DSC$^{10}$) and all (IoU$^{All}$, DSC$^{All}$) activated prototypes with the annotated ROI. For selecting the top-1 and top-10 prototypes, we score the prototypes using the absolute magnitude of the similarity score $\times$ weight of the prototype with the groundtruth class.

\subsection{Visualization of prototypes}
We visualize the local and global explanations of ProtoPNet, BRAIxProtoPNet++ and PIP-Net on CBIS-DDSM and CMMD dataset for qualitative evaluation. We randomly select a test instance for the local explanation and show the top-3 activated prototypes. For global explanation, we show the top-10 image patches for each of eight reasonable looking prototypes. ProtoPNet and PIP-Net have different approaches for visualizing the prototypes. ProtoPNet does not have a fixed patch size. It takes the $x$ percentile from the upsampled similarity map to visualize the patch corresponding to the prototype. PIP-Net has a fixed patch size for prototype visualization. It takes the highest similarity score from the feature map, upsamples it to the patch size in the image space and uses that patch to visualize the prototype. For local visualization, we used the visualization method specific to each model, where we set percentile as 99.9 for ProtoPNet and BraixProtoPNet++ to select the relevant region and for PIP-Net, we set the patch size as $130 \times 130$. However, we used a fixed patch size of $130 \times 130$ for visualization of global explanation for all models. We also used the same fixed patch size for calculation of purity and localization across all models for a fair comparison of the scores.

\section{Results and Discussion}
\label{sec:res}

In this section, we report the classification performance of black-box vs prototype-based models, the qualitative evaluation of the prototypes through local and global explanation and the quantitative evaluation of the prototypes using our prototype evaluation framework for coherence.

\subsection{Performance Comparison of Black-box vs Prototype-based Models}
We compared the classification performance (F1 and AUC) of three intrinsically interpretable prototype-based models (ProtoPNet~\cite{chen2019looks}, BRAIxProtoPNet++~\cite{wang2022knowledge} and PIP-Net~\cite{nauta2023pip}), and three black-box models (EfficientNet~\cite{tan2019efficientnet}, ConvNext~\cite{liu2022convnet} and GMIC~\cite{shen2021interpretable}) (cf. Table~\ref{tab:results:model-results}). ConvNext achieves the highest performance (F1 score) for VinDr (F1: 0.72) and CMMD (F1: 0.74) dataset and GMIC achieves the highest performance (F1: 0.68) for CBIS, with ConvNext having competitive performance (F1: 0.67). However, prototype-based models have competitive performance with black-box models, with BRAIxProtoPNet++ achieving highest AUC (0.86) on CMMD. BRAIxProtoPNet++ outperforms ProtoPNet and PIP-Net in terms of AUC score on all datasets. Note that though PIP-Net has a lower AUC score compared to ProtoPNet and BRAIxProtoPNet++, the F1 score is competitive with the other models even with a much lower number of activated prototypes.
BRAIxProtoPNet++ did not have a public code. We implemented their model and reach a nearby performance to the ones reported in~\cite{wang2023interpretable} on CMMD (BRAIxProtoPNet++ AUC$^{\text{Reported}}$ 88\% vs AUC$^{Ours}$ 86\%; ProtoPNet AUC$^{\text{Reported}}$ 85\% vs AUC$^{Ours}$ 84\%).

\begin{table*}[t!]
\setlength{\tabcolsep}{5pt}
\centering
\caption{Performance comparison of SOTA breast cancer prediction models. Showing mean and standard deviation; values normalized to percentages. Our results show that black-box models outperform prototype-based models, however, prototype-based model are competitive in performance.}
\label{tab:results:model-results}
\begin{small}
\begin{tabular}{+l^l^l^l^l^l^l}
\toprule \tabhead
 & \multicolumn{2}{c}{\textbf{\cbis}} & \multicolumn{2}{c}{\textbf{VinDr}} & \multicolumn{2}{c}{\textbf{CMMD}} \\
& F1 & AUC & F1$^{wM}$ & AUC$^{M}$ & F1 & AUC \\\otoprule
\multicolumn{7}{c}{\textsc{Black box models}}\\
EfficientNet~\cite{tan2019efficientnet} & $58 \pm 6$ & $76 \pm 1$ & $65 \pm 4$ & $77 \pm 1$ & $73 \pm 3$ & $83 \pm 1$ \\
ConvNext~\cite{liu2022convnet} & $67 \pm 3$ & $\textbf{80} \pm 2$ & $\textbf{72} \pm 1$ & $\textbf{83} \pm 0$ & $\textbf{74} \pm 2$ & $84 \pm 1$\\
GMIC~\cite{shen2021interpretable} & $\textbf{68} \pm 3$ & $\textbf{80} \pm 2$ & $64 \pm 3$ & $80 \pm 1$ & $72 \pm 5$ & $83 \pm 1$\\ \midrule
\multicolumn{7}{c}{\textsc{Prototype-based models with ConvNext backbone}}\\
ProtoPNet~\cite{chen2019looks} & $65 \pm 1$ & $77 \pm 1$ & $58 \pm 13$ & $79 \pm 1$ & $70 \pm 0$ & $83 \pm 1$ \\
BRAIxProtoPNet++~\cite{wang2022knowledge} & $64 \pm 4$ & $79 \pm 1$ & $63 \pm 11$ & $81 \pm 0$ & $72 \pm 3$ & $\textbf{86} \pm 2$  \\
PIP-Net~\cite{nauta2023pip} & $63 \pm 3$ & $75 \pm 2$ & $63 \pm 3$ & $78 \pm 1$ & $70 \pm 1$ & $81 \pm 1$ \\
\bottomrule
\end{tabular}
\end{small}
\end{table*}

\subsection{Local and Global Visualization of Prototypes}
In this section, we report some observations from global and local visualization of the prototype-based model. 

\subsubsection{Global explanation}
\begin{figure*}[t!]
    \centering
    \begin{subfigure}[t]{0.32\textwidth}
        \centering
        \includegraphics[scale=0.175]{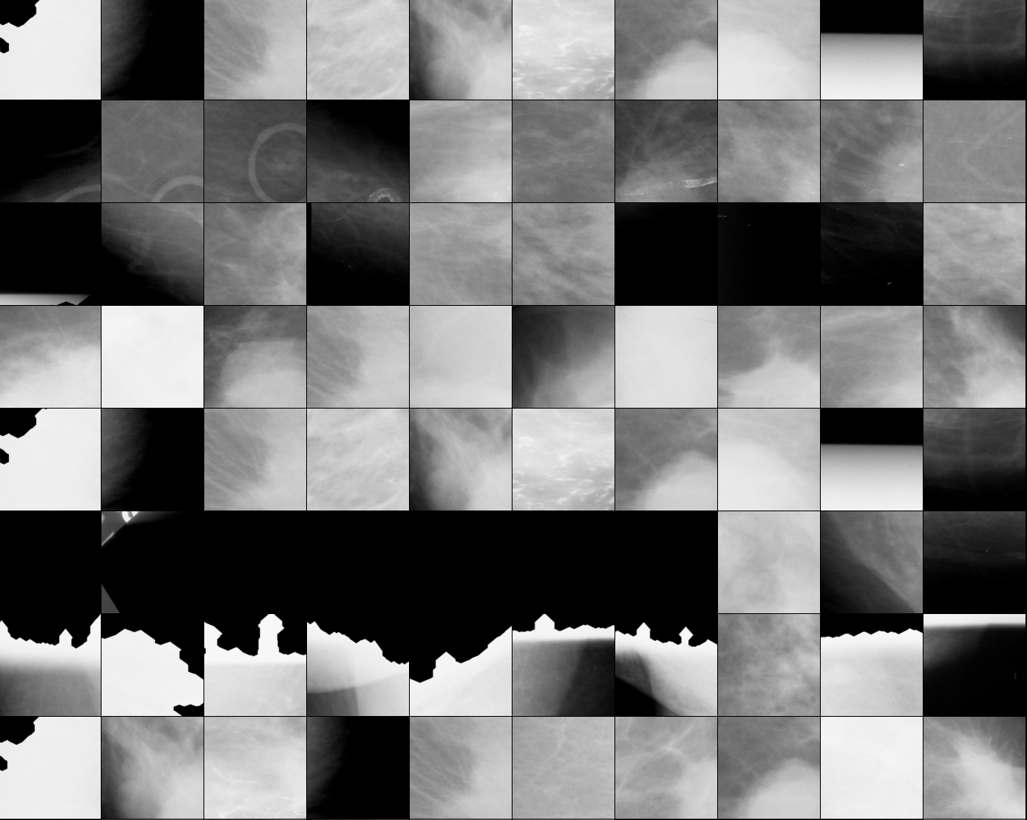}
        \caption{ProtoPNet}
        \label{fig:res:globalexp-cbis-proto}
    \end{subfigure}%
    ~ 
    \begin{subfigure}[t]{0.32\textwidth}
        \centering
        \includegraphics[scale=0.165]{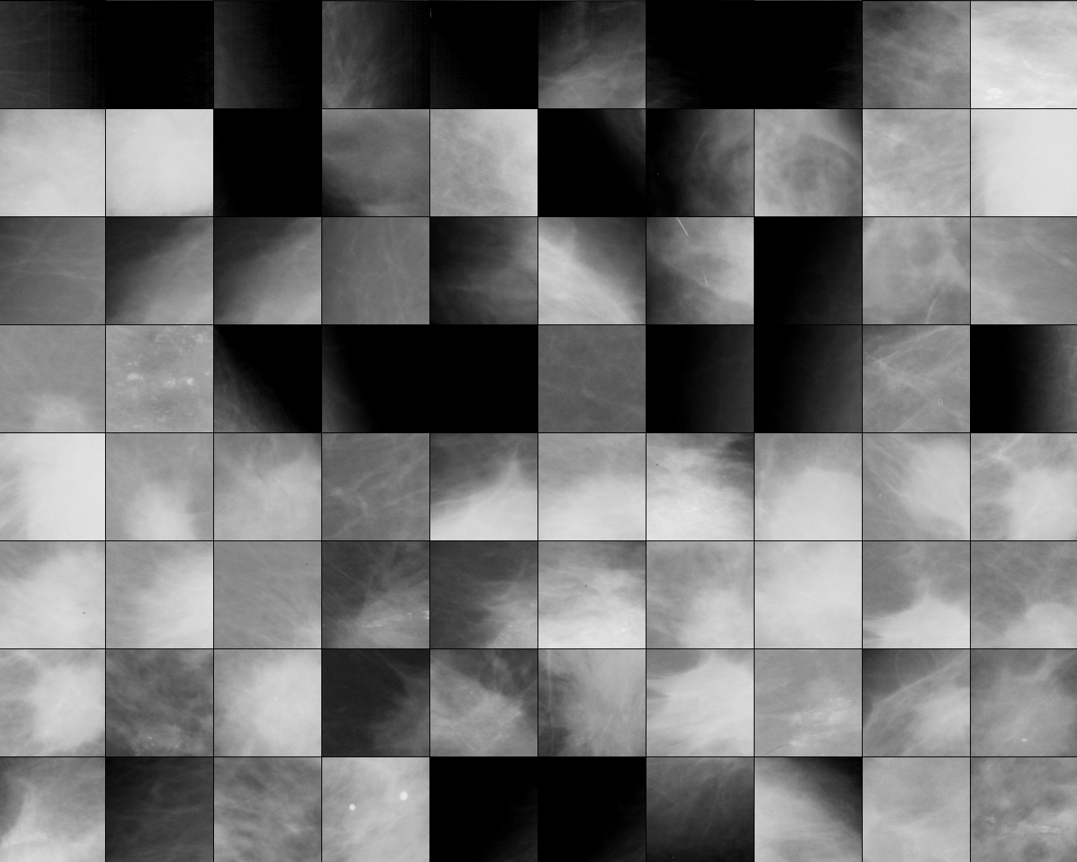}
        \caption{BRAIxProtoPNet++}
        \label{fig:res:globalexp-cbis-braixx}
    \end{subfigure}%
    ~
    \begin{subfigure}[t]{0.32\textwidth}
        \centering
        \includegraphics[scale=0.165]{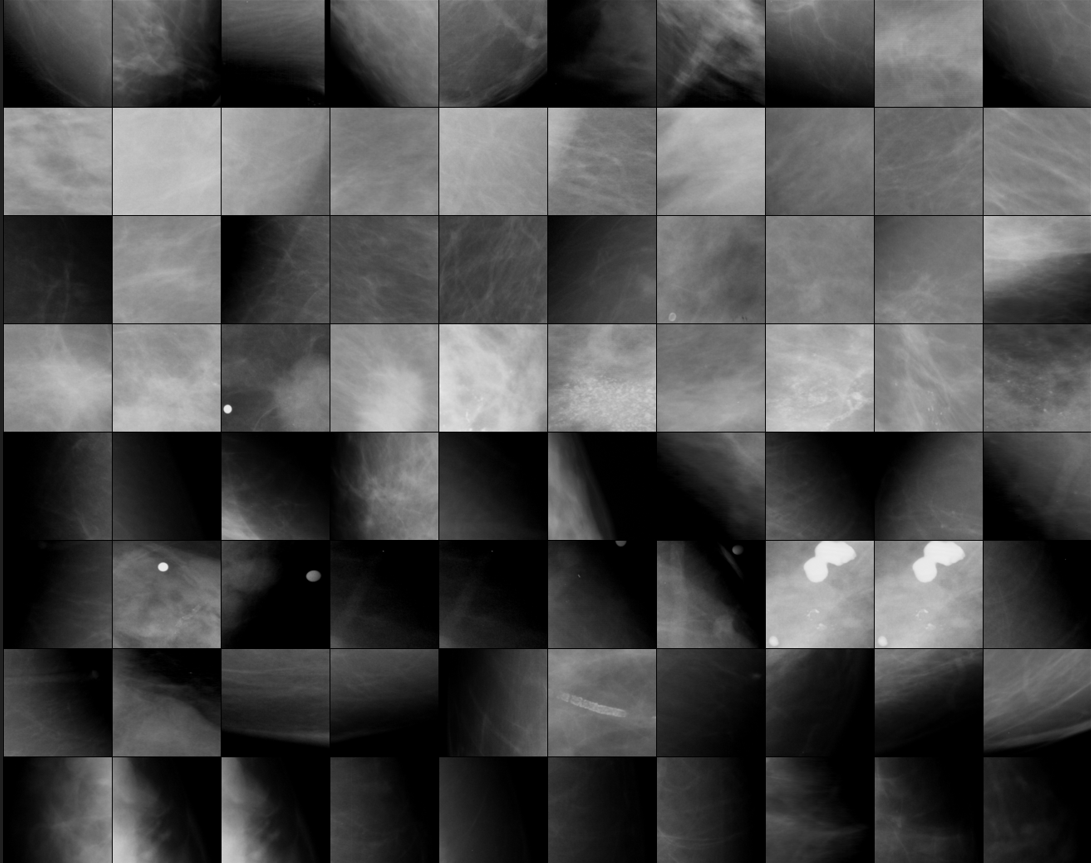}
        \caption{PIP-Net}
        \label{fig:res:globalexp-cbis-pipnet}
    \end{subfigure}
    \caption{Global visualization of 3 prototype-based models trained on CBIS-DDSM dataset. Each row represents one prototype visualized with the top-10 activated image patches from the training set. Example of a prototype description -- second row in ProtoPNet shows mass ROI of irregular shape and spiculated margin.}
    \label{fig:res:globalexp-cbis}
\end{figure*}

\begin{figure*}[t!]
    \centering
    \begin{subfigure}[t]{0.32\textwidth}
        \centering
        \includegraphics[scale=0.18]{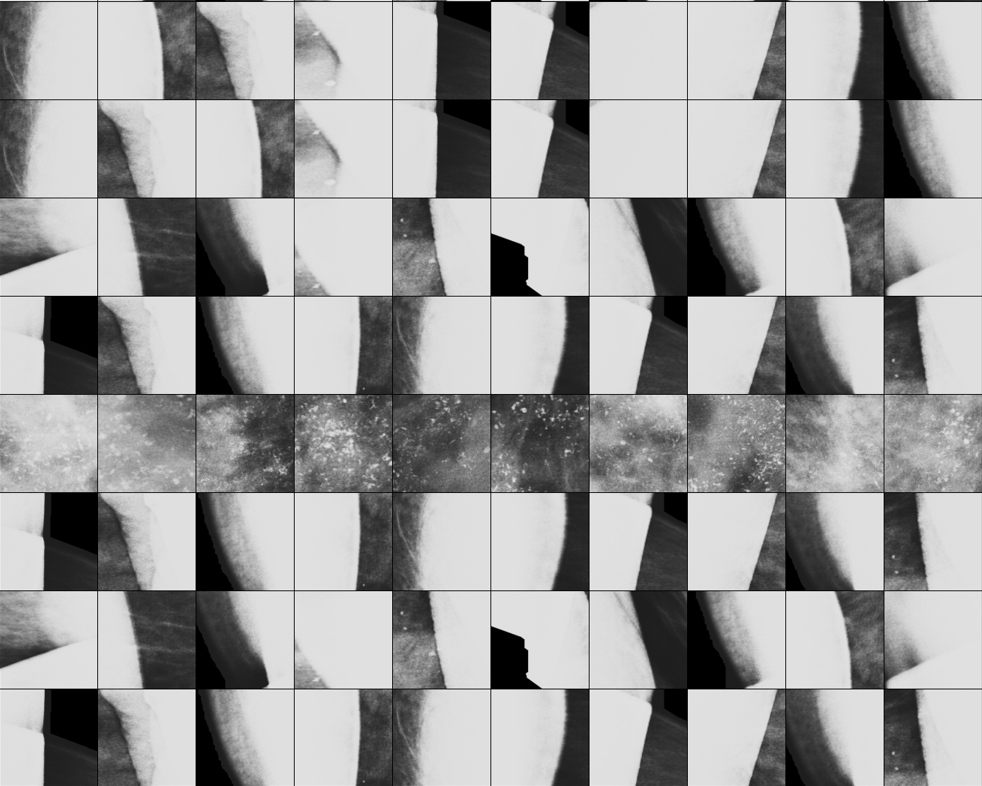}
        \caption{ProtoPNet}
        \label{fig:res:globalexp-cmmd-proto}
    \end{subfigure}%
    ~ 
    \begin{subfigure}[t]{0.32\textwidth}
        \centering
        \includegraphics[scale=0.16]{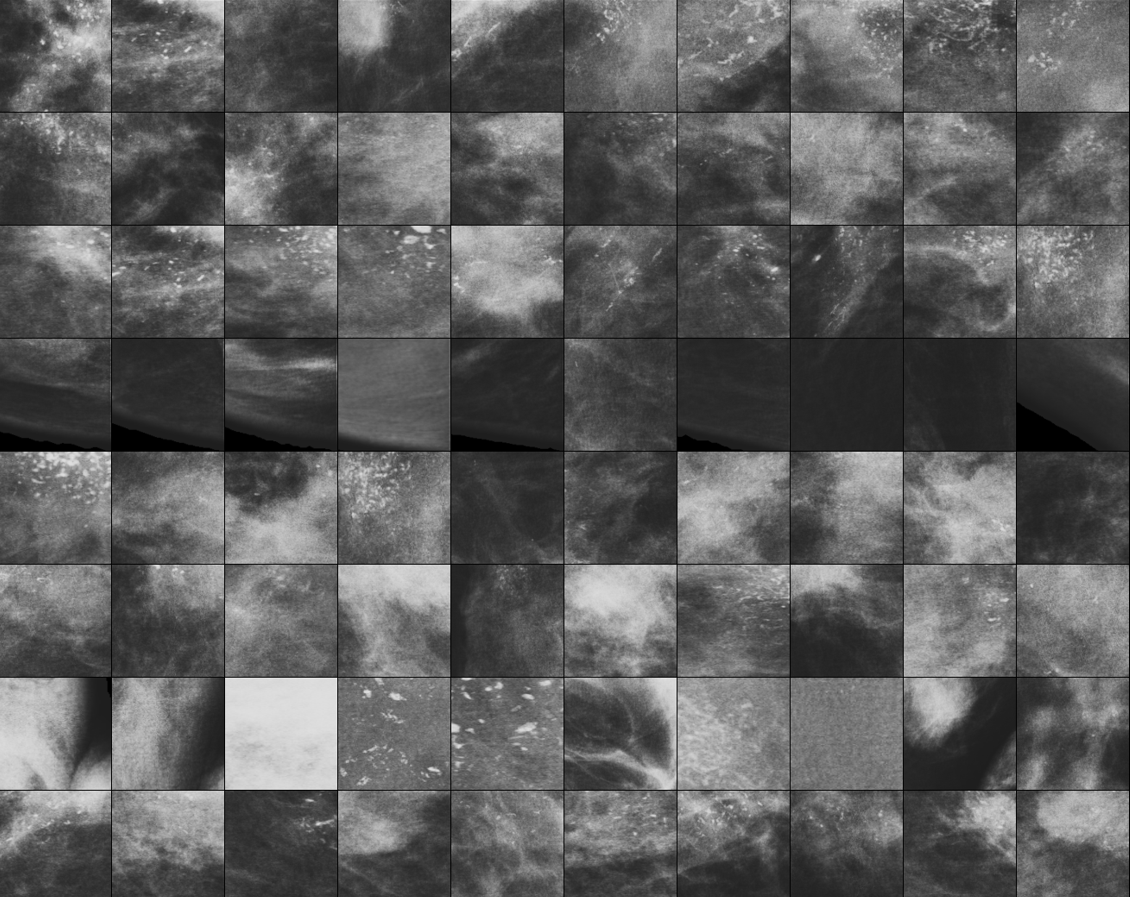}
        \caption{BRAIxProtoPNet++}
        \label{fig:res:globalexp-cmmd-braixx}
    \end{subfigure}%
    ~
    \begin{subfigure}[t]{0.32\textwidth}
        \centering
        \includegraphics[scale=0.17]{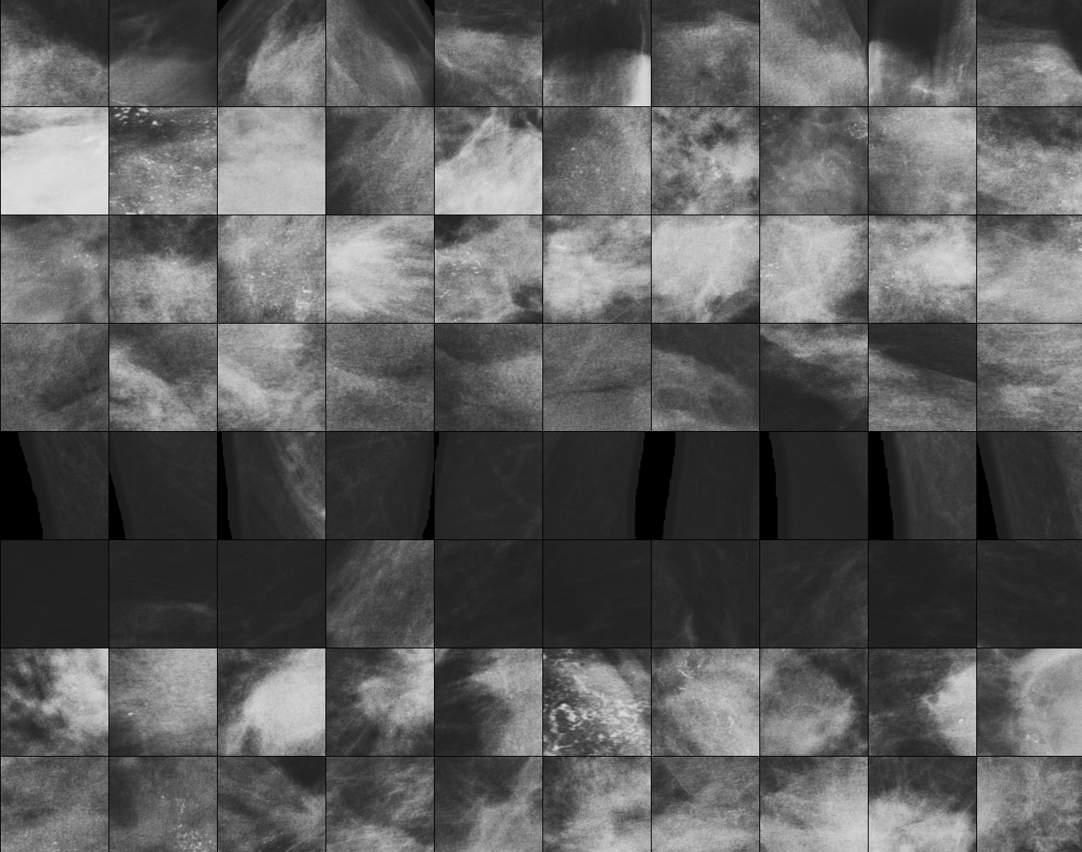}
        \caption{PIP-Net}
        \label{fig:res:globalexp-cmmd-pipnet}
    \end{subfigure}
    \caption{Global visualization of 3 prototype-based models trained on CMMD dataset. Each row represents one prototype visualized with the top-10 activated image patches from the training set. Example of a prototype description - fifth row in ProtoPNet shows calcification abnormality.}
    \label{fig:res:globalexp-cmmd}
\end{figure*}

We visualized the top-10 image patches (based on the highest similarity score) activated by each prototype for the three prototype-based models for CBIS-DDSM (cf. Fig.~\ref{fig:res:globalexp-cbis}) and CMMD dataset (cf. Fig~\ref{fig:res:globalexp-cmmd}). 

\textbf{Not all top-k patches activated by a prototype are visually similar.} We can see prototypes representing the mass abnormality in the visualization from CBIS-DDSM. In ProtoPNet global explanation (cf. Fig~\ref{fig:res:globalexp-cbis-proto}), the prototype in the second row contains some patches with mass ROI of irregular shape and spiculated margin and the fourth row contains some mass ROIs of oval shape and ill-defined margin, among other different looking patched showing normal tissue. This shows that all patches activated by a prototype in ProtoPNet may not be visually similar. Also, the same prototype can get activated for different abnormality categories. For example, in Fig.~\ref{fig:res:globalexp-cbis-pipnet}, prototype in 4th row contains most ROIs of category mass with irregular shape and spiculated margin. However, that prototype also contains ROIs of category calcification with pleomorphic and amorphous morphology and segmental distribution. 


\textbf{Duplicate prototypes learnt.} Separate prototypes may learn the same abnormality category as shown in Fig~\ref{fig:res:globalexp-cbis-braixx}, where prototypes in 5th, 6th and 7th row all contain mass ROIs of irregular shape and spiculated margin. In Fig.~\ref{fig:res:globalexp-cmmd-proto}, it can be seen that ProtoPNet has learnt duplicate prototypes of pectoral muscle (a triangular region seen on one side of the image).

\textbf{Irrelevant prototypes learnt.} There are prototypes representing normal tissue and black background in all prototype-based models. This is expected, however, such prototypes should have lower weights to the classes as they usually do not have much contribution to benign and malignant prediction. 

Overall, it can be observed that i) duplicate prototypes might be learnt by the models (quite easily visually observed in ProtoPNet (cf. Fig~\ref{fig:res:globalexp-cmmd-proto})), ii) all top-10 patches activated by one prototype may not look visually the same or may not belong to the same abnormality category, and iii) not all learnt prototypes belong to the ROIs, but can belong to edges and background having non-zero weights in the classification layer. 

\begin{figure*}[thbp]
\centering
    \begin{subfigure}[b]{\textwidth}
        \includegraphics[scale=0.30]{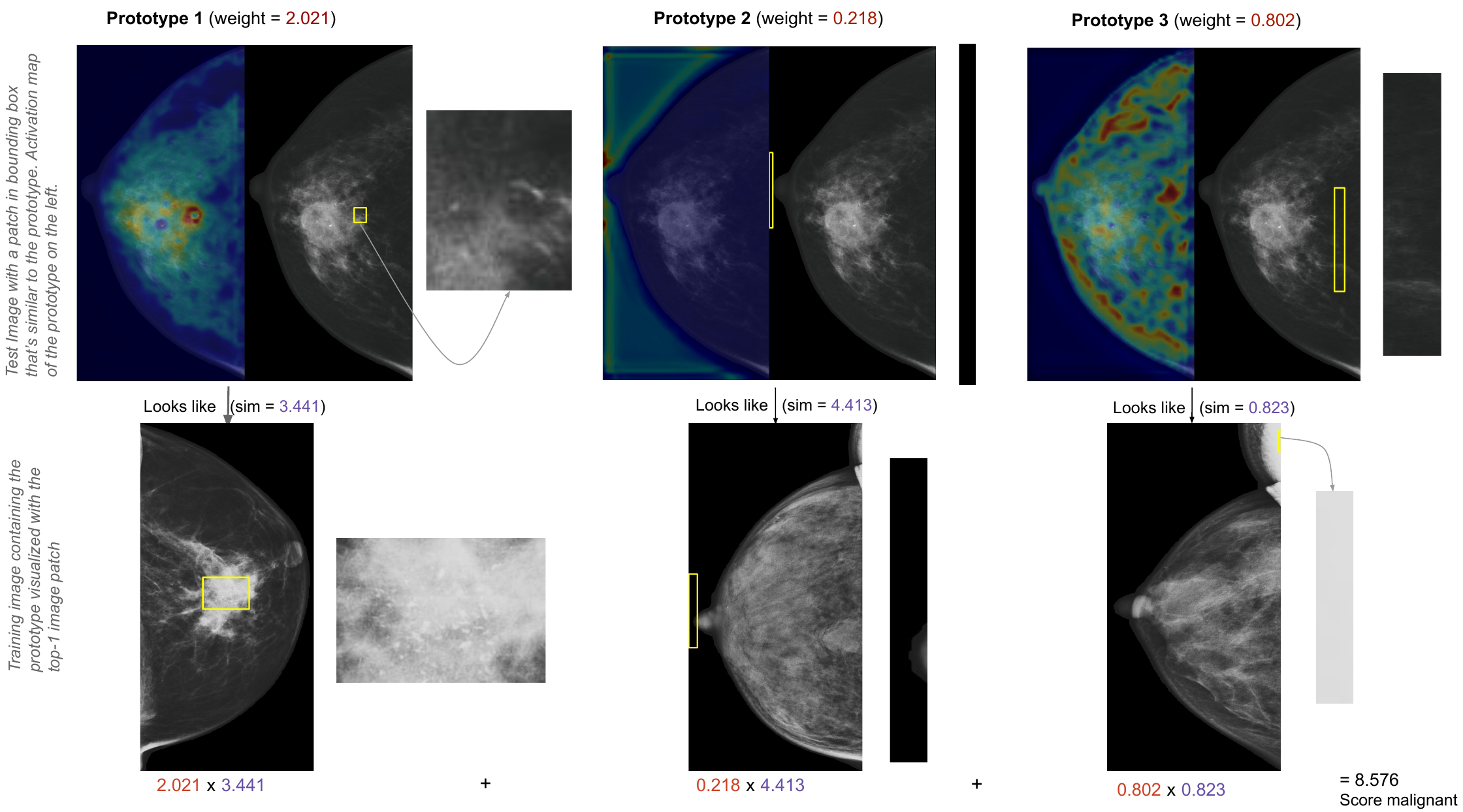}
        \caption{Prototypes activated for malignant class}
    \end{subfigure}
    \begin{subfigure}[b]{\textwidth}
        \includegraphics[scale=0.30]{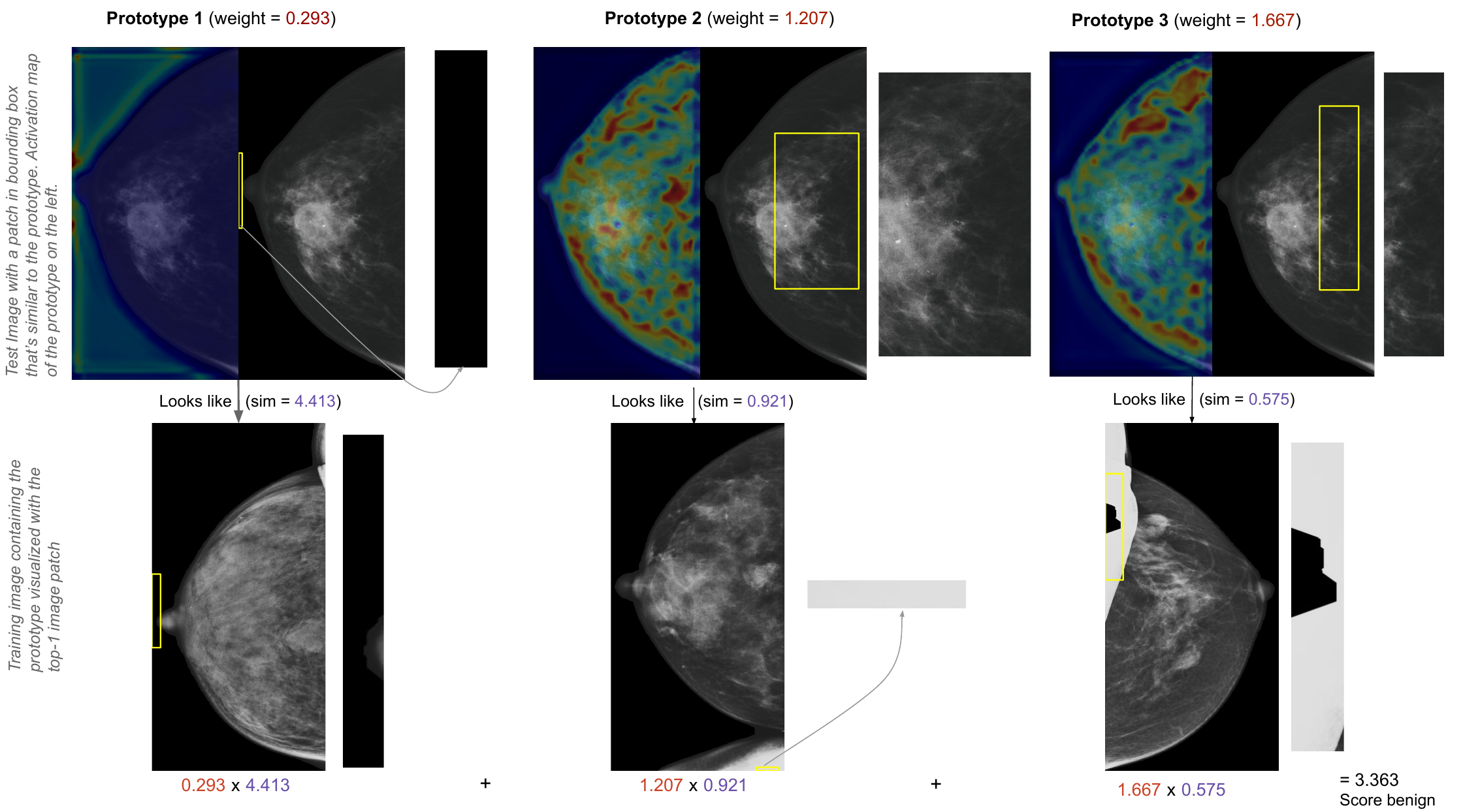}
        \caption{Prototypes activated for benign class}
    \end{subfigure}
\caption{Local explanation from ProtoPNet showing the top-3 activated prototypes for the malignant class and the benign class. Example image: CMMD, malignant test case D2-0249, view RCC, predicted class malignant.}
\label{fig:res:protopnet-localexp-cmmd}
\end{figure*}

\begin{figure*}[thbp]
\centering
    \begin{subfigure}[b]{\textwidth}
        \includegraphics[scale=0.30]{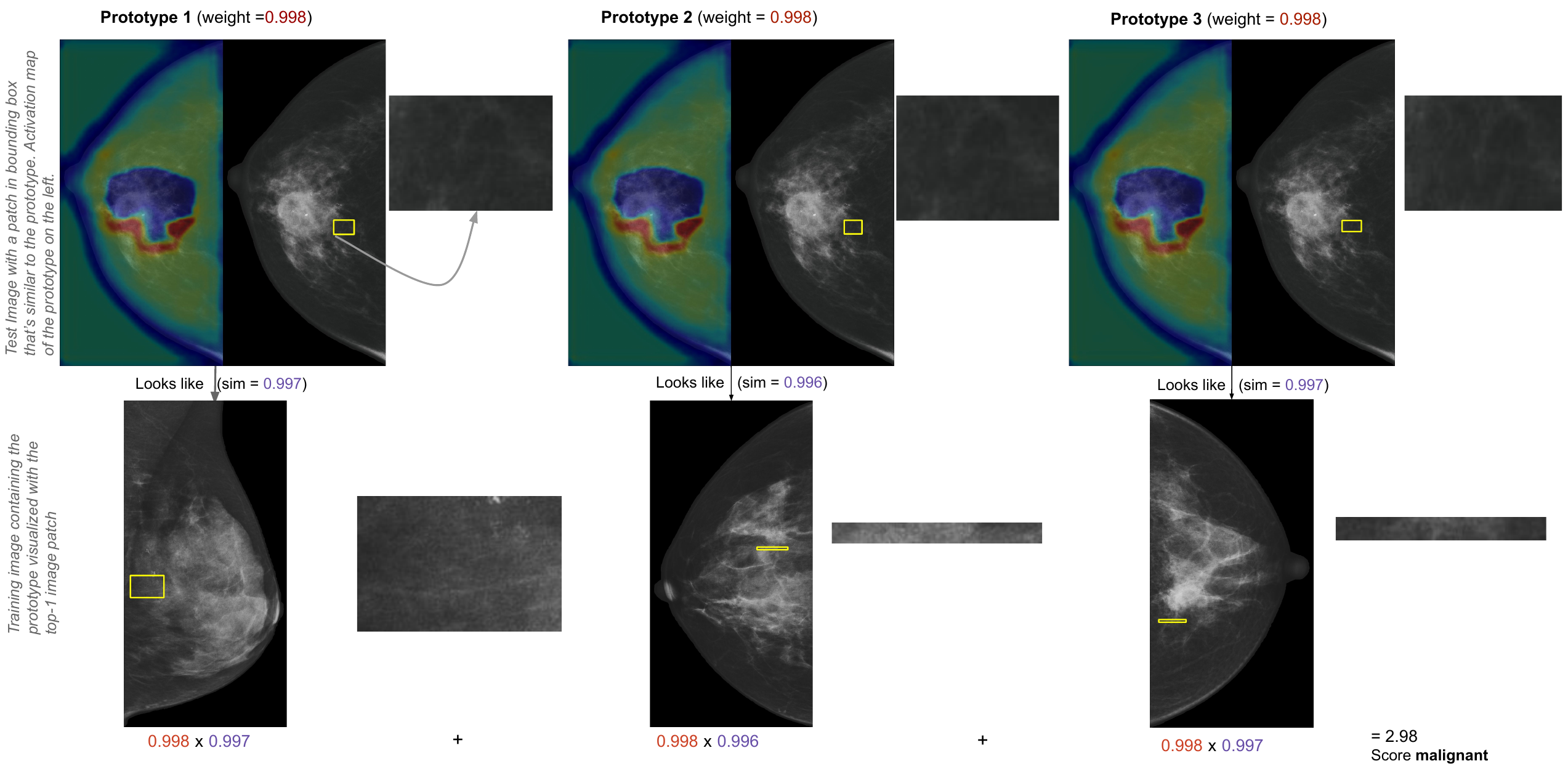}
        \caption{Prototypes activated for malignant class}
    \end{subfigure}
    \begin{subfigure}[b]{\textwidth}
        \includegraphics[scale=0.30]{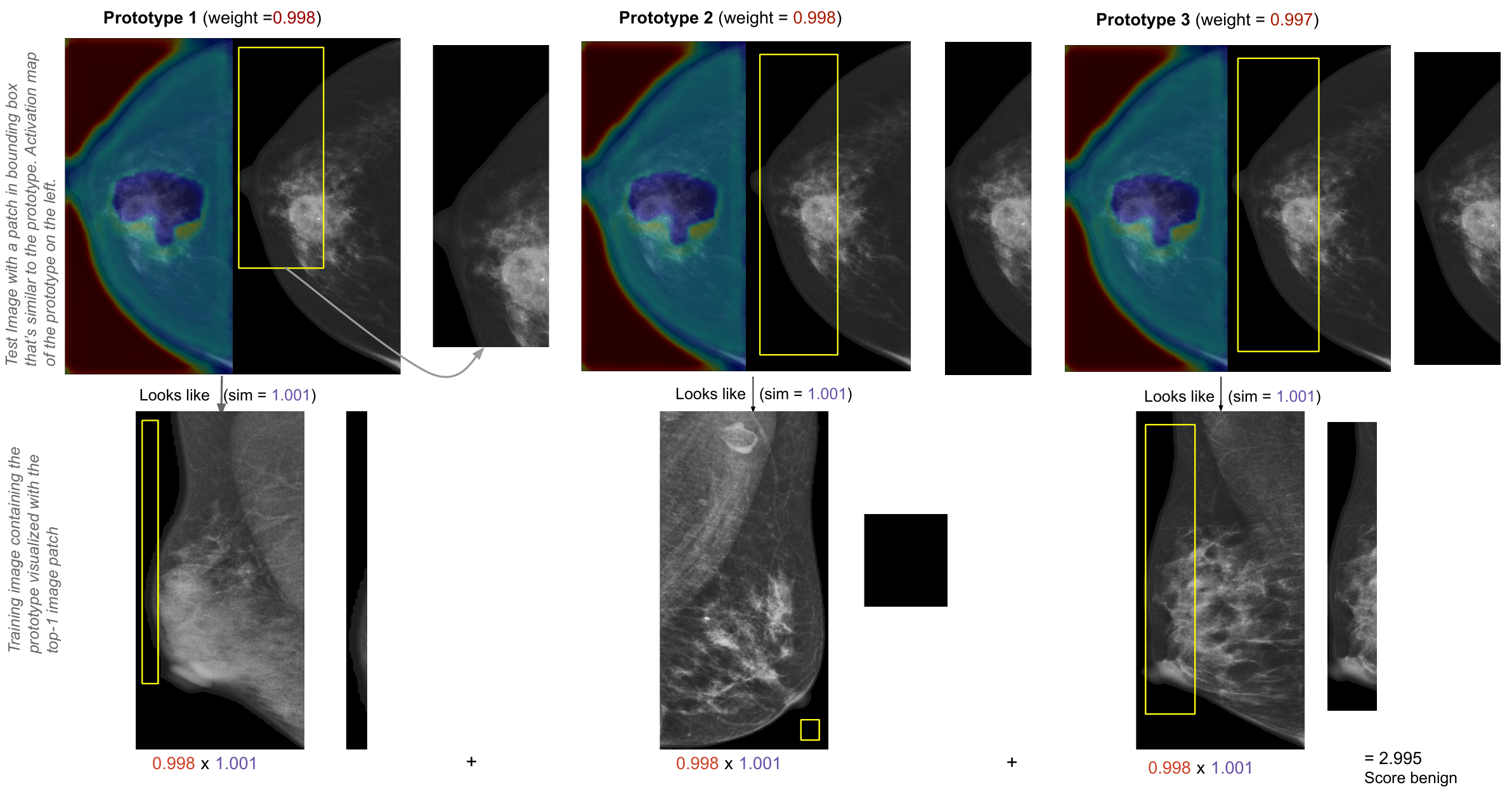}
        \caption{Prototypes activated for benign class}
    \end{subfigure}
\caption{Local explanation from BRAIxProtoPNet++ showing the top-3 activated prototypes for the malignant class and the benign class. Example image: CMMD, malignant test case D2-0249, view RCC, predicted class malignant.}
\label{fig:res:braixx-localexp-cmmd}
\end{figure*}

\begin{figure*}[thbp]
\centering
    \begin{subfigure}[b]{\textwidth}
        \includegraphics[scale=0.35]{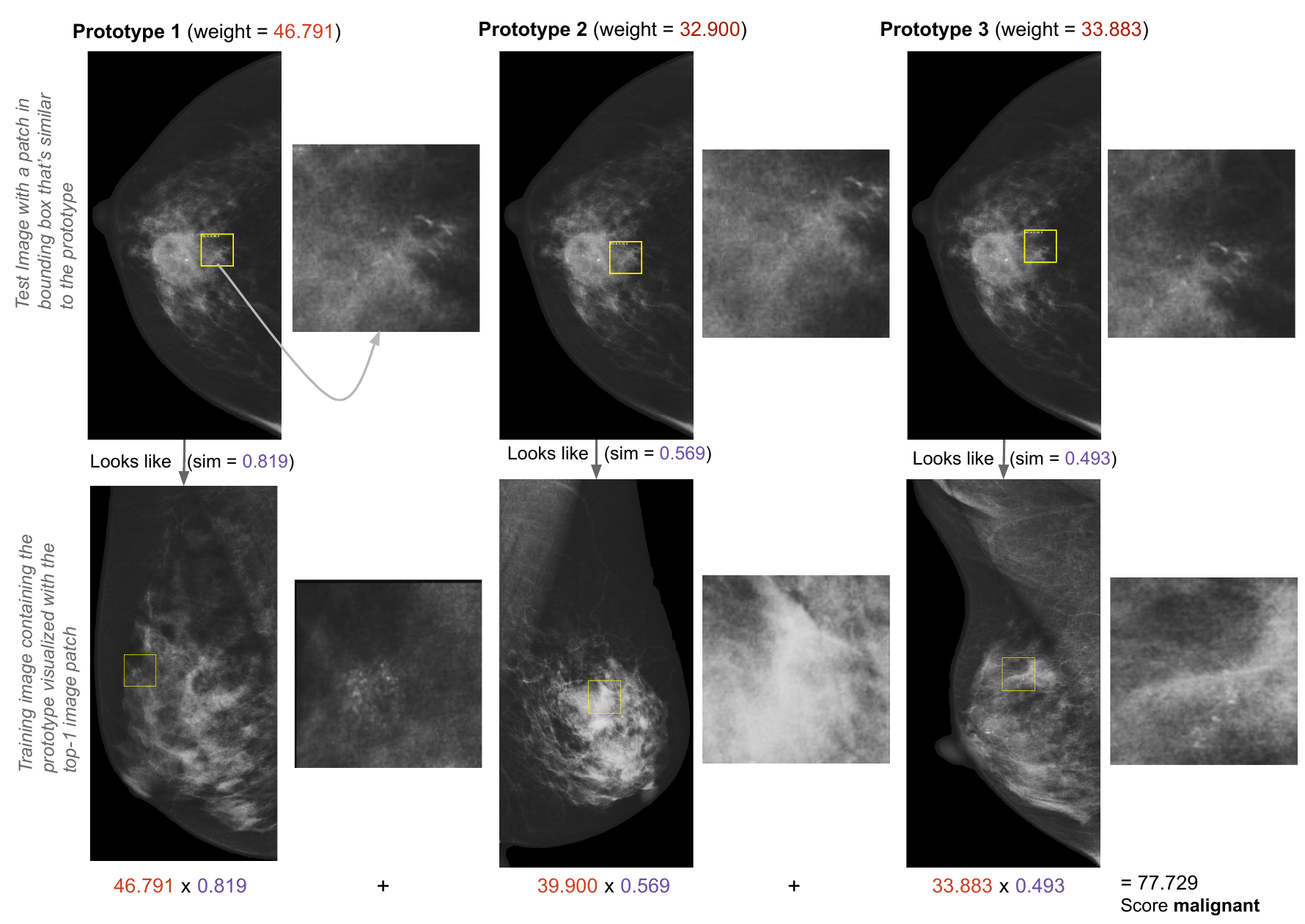}
        \caption{Prototypes activated for malignant class}
    \end{subfigure}
    \begin{subfigure}[b]{\textwidth}
        \includegraphics[scale=0.35]{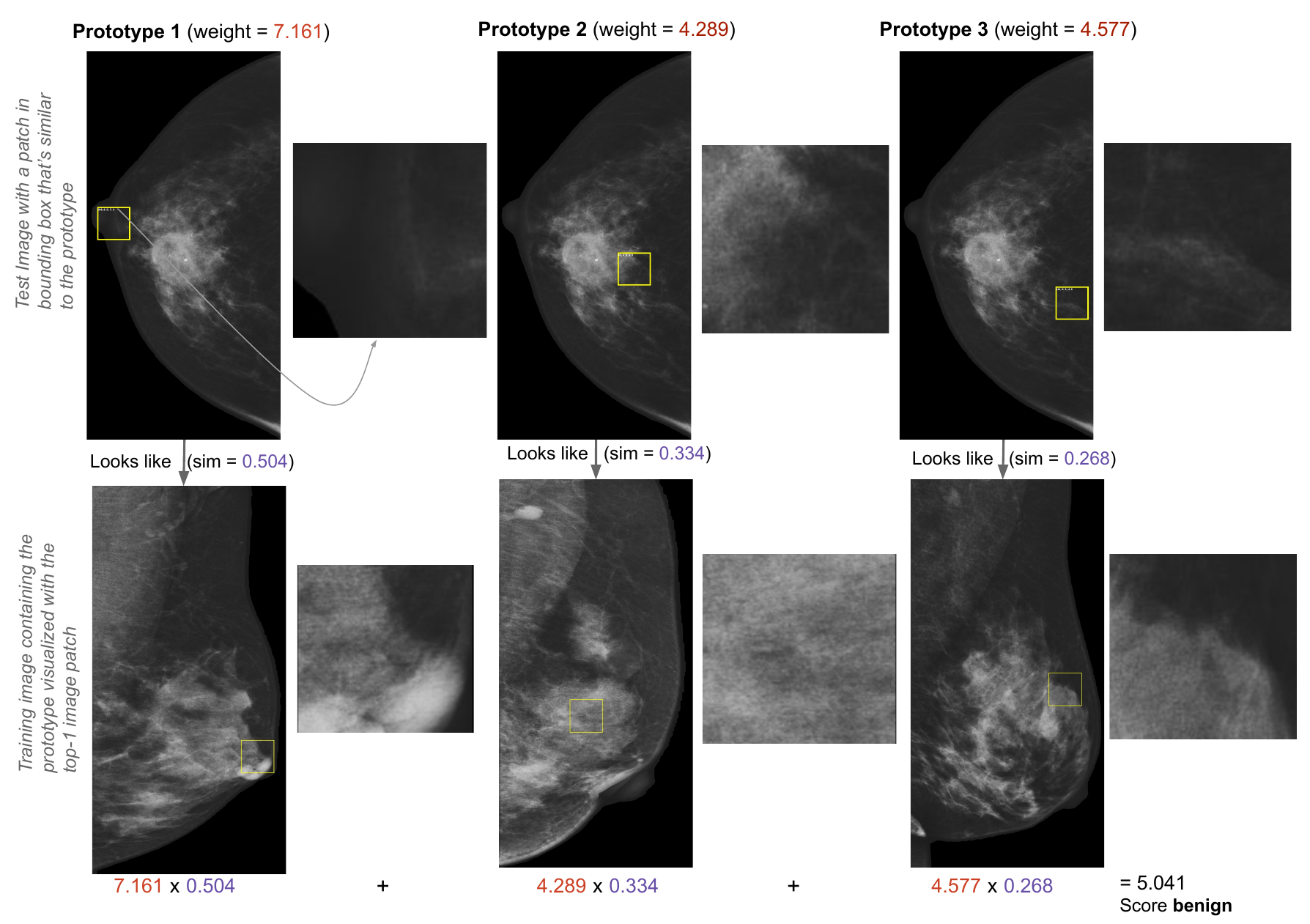}
        \caption{Prototypes activated for benign class}
    \end{subfigure}
\caption{Local explanation from PIP-Net showing the top-3 activated prototypes for the malignant class and the benign class. Example image: CMMD, malignant test case D2-0249, view RCC, predicted class malignant.}
\label{fig:res:pipnet-localexp-cmmd}
\end{figure*}

\subsubsection{Local explanation}
We visualized local explanation for one test instance from CMMD dataset for ProtoPNet (cf. Fig.~\ref{fig:res:protopnet-localexp-cmmd}), BRAIxProtoPNet++ (cf. Fig.~\ref{fig:res:braixx-localexp-cmmd}) and PIP-Net (cf. Fig.~\ref{fig:res:pipnet-localexp-cmmd}) as an anecdotal evidence. We show the top-3 activated prototypes for each model. 

For ProtoPNet (cf. Fig.~\ref{fig:res:protopnet-localexp-cmmd}), we observed that i) one of the activated prototypes contained the correct ROI, ii) however, the top activated prototypes also represented some irrelevant black region and edges, iii) the test image patch does not always look similar to the prototype from the training set, and iv) similar looking prototype may get activated for both benign and malignant classes. 

For BRAIxProtoPNet++ (cf. Fig.~\ref{fig:res:braixx-localexp-cmmd}), we observed that i) the region surrounding the actual ROI gets activated rather than the actual ROI, ii) most of the activated prototypes point to the same region (we have only shown for 3, but we found other prototypes to also point to the same region as the top 3), and iii) the activated prototypes of the benign class (not predicted class) represent the background. 

For PIP-Net, we observed that i) the top activated prototypes contain the ROIs, ii) the test image patch has some similarity to the prototype from the training set, and iii) the benign prototypes don't belong to the ROI, but to the other regions of the breast. 

Overall, the local explanation of PIP-Net for this test image looked more reasonable compared to the other models as the top-3 prototypes located the correct region for predicting the correct class. 

\begin{table}[th!bp]
\centering
\setlength{\tabcolsep}{5pt}
\caption{Comparison of prototype-based models using our prototype evaluation framework on CBIS-DDSM dataset. Mean and standard deviation reported across three model runs with different seeds. Local prototypes show the number of local prototypes active with positive and negative scores (similarity $\times$ weight). Sparsity ratio is the percentage of total prototypes that have zero weights to all classes in the classification layer. Our results show that prototype-based models for breast cancer prediction task need to improve on i) reducing irrelevant prototypes, ii) purity of the prototypes, iii) learning more unique prototypes such
that the prototypes can cover more abnormality categories, and iv) finding the ROIs more accurately. } %
\label{tab:results:pef-c}
\begin{threeparttable}
\begin{small}
\begin{tabular}{+l^c^c^c} 
\toprule \tabhead
Property & ProtoPNet & BRAIxProtoPNet++ & PIP-Net  \\ \otoprule
\textit{Compactness} & & & \\
\hspace{0.5cm}Global $\downarrow$ & $400 \pm 0.0$ & $400 \pm 0.0$ & $90 \pm 83$\\ 
\hspace{0.5cm}Local $\downarrow$ & & & \\
\hspace{0.8cm}Positive & $314 \pm 20$ & $200 \pm 0$ & $27 \pm 31$\\
\hspace{0.8cm}Negative & $84 \pm 20$ & $200 \pm 0$ & - \\
\hspace{0.5cm}Sparsity $\uparrow$ & 0\% & 0\% & 88\% \\
\textit{Relevance} $\uparrow$ & $\textbf{0.36} \pm 0.23$ & $0.30 \pm 0.07$ & $0.28 \pm 0.05$ \\
\textit{Specialization} $\uparrow$ & & & \\
\hspace{0.5cm}Abnorm. Type & $\textbf{0.37} \pm 0.02$ & $0.22 \pm 0.09$ & $0.31 \pm 0.13$\\
\hspace{0.5cm}Mass Shape & $0.29 \pm 0.04$ & $0.29 \pm 0.07$ & $\textbf{0.35} \pm 0.08$\\
\hspace{0.5cm}Mass Margin & $0.34 \pm 0.03$ & $0.33 \pm 0.10$ & $\textbf{0.38} \pm 0.04$\\
\hspace{0.5cm}Calc. Morph. & $0.30 \pm 0.22$ & $0.24 \pm 0.06$ & $\textbf{0.32} \pm 0.09$\\
\hspace{0.5cm}Calc. Distribution & $0.24 \pm 0.11$ & $0.24 \pm 0.03$ & $\textbf{0.27} \pm 0.09$\\
\textit{Uniqueness} $\uparrow$ & $0.13 \pm 0.07$ & $0.31 \pm 0.08$ & $\textbf{0.53} \pm 0.12$\\ 
\textit{Coverage} $\uparrow$ & $0.12 \pm 0.05$ & $\textbf{0.25} \pm 0.02$ & $0.10 \pm 0.08$\\
\textit{Class-specific} $\uparrow$ & $0.52 \pm 0.05$ & $0.61 \pm 0.03$ & $\textbf{0.65} \pm 0.08$\\
\textit{Localization} $\uparrow$ & & & \\
\hspace{0.5cm}IoU$^{1}$ & $0.04 \pm 0.01$ & $0.04 \pm 0.02$ & $\textbf{0.07} \pm 0.00$\\
\hspace{0.5cm}IoU$^{10}$ & $\textbf{0.13} \pm 0.02$ & $0.10 \pm 0.01$ & $\textbf{0.13} \pm 0.01$\\
\hspace{0.5cm}IoU$^{All}$ & $0.23 \pm 0.03$ & $\textbf{0.28} \pm 0.01$ & $0.18 \pm 0.06$\\
\hspace{0.5cm}DSC$^{1}$ & $0.06 \pm 0.02$ & $0.06 \pm 0.03$ & $\textbf{0.10} \pm 0.0$ \\
\hspace{0.5cm}DSC$^{10}$ & $\textbf{0.20} \pm 0.03$ & $0.15 \pm 0.02$ & $0.19 \pm 0.01$\\
\hspace{0.5cm}DSC$^{All}$ & $0.34 \pm 0.04$ & $\textbf{0.39} \pm 0.02$ & $0.26 \pm 0.09$\\
\bottomrule
\end{tabular}
\begin{tablenotes}
Note: The black-box model, GMIC has an IoU of $0.05 \pm 0.0$ and DSC of $0.09 \pm 0.0$ on CBIS-DDSM over 6 ROI candidates that are extracted from the GMIC model. 
\end{tablenotes}
\end{small}
\end{threeparttable}
\end{table}

\subsection{Automatic Quantitative Evaluation of Prototypes}
We report the results of our prototype evaluation framework in Table~\ref{tab:results:pef-c} for ProtoPNet, BRAIxProtoPNet++ and PIP-Net on CBIS-DDSM dataset. 

\textbf{PIP-Net is more compact.} We observe that PIP-Net has much lower number of global and local prototypes increasing the sparsity of the model.  

\textbf{Majority of the global prototypes are irrelevant.} ProtoPNet has higher \textit{Relevance} score compared to the others suggesting more prototypes get activated on ROIs, however, the standard deviation is quite high reducing the reliability of the score. Relevance score ranges from 0.28 to 0.36 for the 3 models, showing that majority of the global prototypes are not associated with a ROI. 

\textbf{Relevant prototypes are not fully pure.} \textit{Specialization} scores show that ProtoPNet has the highest purity at the first granularity level of abnormality type, however, PIP-Net has the highest purity in the second granularity level. This shows that though ProtoPNet is better at learning prototypes representing only mass or calcification, PIP-Net is better at learning prototypes representing a particular type of mass or calcification, which is in line with the `semantic correspondence' characteristic of the model. 

\textbf{More varied relevant prototypes need to be learnt.} PIP-Net has the highest \textit{Uniqueness} score suggesting that most of its relevant prototypes belong to a unique category. However, the high uniqueness score is also an effect of the low global prototypes. For example, in one run of PIP-Net, 16 out of 48 global prototypes were relevant and of these 16, 10 were unique, resulting in a high uniqueness score of 0.625. Whereas in one run of BRAIxProtoPNet++, 135 out of 400 global prototypes were relevant, of which 33 were unique resulting in a lower uniqueness score of 0.24. Therefore, it is better to look at the \textit{Uniqueness} measure together with the \textit{Coverage} measure, which shows BRAIxProtoPNet++ to have a higher coverage than PIP-Net due to higher unique categories. 

\textbf{Most of the relevant prototypes are associated with the correct class.} We also measured \textit{Class-specific} score, i.e. how much do the class weights of the relevant prototypes (associated with a category) align with the class distribution of that category from the dataset. PIP-Net has the highest class-specific score suggesting that more relevant prototypes (representing the ROIs) are associated the correct class compared to the other models. However, PIP-Net also has a low number of relevant prototypes, compared to the others, which can result in a higher class-specific score. 

\textbf{Localization of the regions-of-interest needs to be improved.} \textit{Localization} score shows that PIP-Net is better in finding the ROI with the top-1 activated prototype. At top-10 activated prototypes, ProtoPNet has similar IoU to PIP-Net and with all activated prototypes, BRAIxProtoPNet++ has the highest IoU due to its high number of global prototypes (400) compared to PIP-Net (90). The IoU$^6$ score of the black-box model, GMIC, is similar to the IoU$^1$ of the prototype based models and with only 10 activated prototypes, the prototype based models outperform GMIC is finding the correct ROI. This shows that prototype-based models can be a good candidate for unsupervised ROI extraction.

\textbf{Summary.} 
Overall, 28-36\% of the global prototypes across the 3 models are relevant. Prototypes of ProtoPNet have the highest purity score of 37\% at the first granularity level of mass and calcification. Prototypes from PIP-Net has the highest purity score of 27-35\% at the second granularity level of a specific type of mass or calcification. 
Prototypes of BRAIxProtoPNet++ cover around 25\% of the total abnormality categories (categories from the BI-RADS lexicon) from the dataset. Around 52-65\% of the relevant prototypes are associated with the correct class across all models. Further, prototype-based model outperformed the black-box model, GMIC, in localizing the ROIs, suggesting that prototype-based models can be a good candidate for unsupervised regions-of-interest extraction. However, overall, the intersection over union scores for top-10 activated prototypes are quite low (around 0.13 for PIP-Net). This suggests that we still need to improve on learning relevant prototypes which can get activated on all ROIs. We found slightly higher scores for PIP-Net w.r.t some properties, but PIP-Net also has a much lower number of global prototypes, which can influence some of the measures. Having lower number of prototypes (compactness) makes the explanation easier for users to comprehend, but we need to increase the number of unique and relevant prototypes learnt by the model to improve the model's detection of various ROIs.

\section{Conclusion and Future Work}
We extensively compare three state-of-the-art black-box models to three state-of-the-art prototype-based models in the context of breast cancer prediction using mammography. We found that though black-box model, ConvNext, has higher classification performance compared to prototype-based models, prototype-based models are competitive in performance. To go beyond anecdotal evidence in evaluating the quality of the prototypes, we propose a \textit{prototype evaluation framework for coherence (PEF-Coh)} to quantitatively evaluate the prototype quality based on domain knowledge. Our framework needs regions-of-interest (ROIs) annotation and some fine-grained labels for automatic evaluation of prototype quality. We used such annotations from a public dataset, CBIS-DDSM and evaluated our framework on three prototype-based models. Our analysis shows that prototype-based models still need to improve on i) reducing irrelevant prototypes, ii) purity of the prototypes, iii) learning more unique prototypes such that the prototypes can cover more abnormality categories and iv) finding the ROIs more accurately. 

In the future, it would be interesting to perform user evaluation of the learnt prototypes to assess their quality with domain-experts. Further, previous re-
search has shown that patch visualization method of ProtoPNet can be imprecise~\cite{gautam2023looks,xu2023sanity} and PRP~\cite{gautam2023looks} visualization method can generate more precise visualization. Further, receptive field of the prototype activated patch may be effected by image regions outside of the visualized region, leading to spatial explanation misalignment~\cite{sacha2024interpretability}. In this work, we have used the standard visualization method of each method reported in their original work. In the future, it would be interesting to investigate the effect of visualization method on the prototype quality.

Our vision is to integrate the interpretable model in the clinical workflow for the purpose of assisting the clinicians with diagnosis. We envision a two-phase approach: a bootstrap-phase where clinicians study the prototypes of the global explanation and discuss their meaning giving them a name and a description; and a use-and-tune phase where clinicians during their everyday diagnosis see and amend the local explanation along with the prototype names and descriptions. In this way, the documentation of the meaning of the prototypes and thereby the quality of the model can gradually improve in a natural manner.

\subsubsection{Disclosure of Interests.}
The authors have no competing interests to declare that are
relevant to the content of this article.

\bibliographystyle{splncs04}
\bibliography{main}

\begin{thebibliography}{10}
\providecommand{\url}[1]{\texttt{#1}}
\providecommand{\urlprefix}{URL }
\providecommand{\doi}[1]{https://doi.org/#1}

\bibitem{barnett2021case}
Barnett, A.J., Schwartz, F.R., Tao, C., Chen, C., Ren, Y., Lo, J.Y., Rudin, C.:
  A case-based interpretable deep learning model for classification of mass
  lesions in digital mammography. Nature Machine Intelligence  \textbf{3}(12),
  1061--1070 (2021)

\bibitem{chen2019looks}
Chen, C., Li, O., Tao, D., Barnett, A., Rudin, C., Su, J.K.: This looks like
  that: deep learning for interpretable image recognition. Advances in neural
  information processing systems  \textbf{32} (2019)

\bibitem{cmmd}
Cui, C., Li, L., Cai, H., Fan, Z., Zhang, L., Dan, T., Li, J., Wang, J.: The
  chinese mammography database (cmmd): An online mammography database with
  biopsy confirmed types for machine diagnosis of breast. (version 1) [data
  set] (2021), \url{https://doi.org/10.7937/tcia.eqde-4b16}, the Cancer Imaging
  Archive, Accessed: 08/09/2023

\bibitem{gautam2023looks}
Gautam, S., H{\"o}hne, M.M.C., Hansen, S., Jenssen, R., Kampffmeyer, M.: This
  looks more like that: Enhancing self-explaining models by prototypical
  relevance propagation. Pattern Recognition  \textbf{136},  109172 (2023)

\bibitem{kim2021xprotonet}
Kim, E., Kim, S., Seo, M., Yoon, S.: Xprotonet: diagnosis in chest radiography
  with global and local explanations. In: Proceedings of the IEEE/CVF
  conference on computer vision and pattern recognition. pp. 15719--15728
  (2021)

\bibitem{liu2022convnet}
Liu, Z., Mao, H., Wu, C.Y., Feichtenhofer, C., Darrell, T., Xie, S.: A convnet
  for the 2020s. In: Proceedings of the IEEE/CVF conference on computer vision
  and pattern recognition. pp. 11976--11986 (2022)

\bibitem{nauta2023pip}
Nauta, M., Schl{\"o}tterer, J., van Keulen, M., Seifert, C.: Pip-net:
  Patch-based intuitive prototypes for interpretable image classification. In:
  Proceedings of the IEEE/CVF Conference on Computer Vision and Pattern
  Recognition. pp. 2744--2753 (2023)

\bibitem{nauta2023co}
Nauta, M., Seifert, C.: The co-12 recipe for evaluating interpretable
  part-prototype image classifiers. In: World Conference on Explainable
  Artificial Intelligence. pp. 397--420. Springer (2023)

\bibitem{nauta2023anecdotal}
Nauta, M., Trienes, J., Pathak, S., Nguyen, E., Peters, M., Schmitt, Y.,
  Schl{\"o}tterer, J., van Keulen, M., Seifert, C.: From anecdotal evidence to
  quantitative evaluation methods: A systematic review on evaluating
  explainable ai. ACM Computing Surveys  \textbf{55}(13s),  1--42 (2023)

\bibitem{nauta2021neural}
Nauta, M., Van~Bree, R., Seifert, C.: Neural prototype trees for interpretable
  fine-grained image recognition. In: Proceedings of the IEEE/CVF Conference on
  Computer Vision and Pattern Recognition. pp. 14933--14943 (2021)

\bibitem{Nguyenvindr}
Nguyen, H.T., Nguyen, H.Q., Pham, H.H., Lam, K., Le, L.T., Dao, M., Vu, V.:
  Vindr-mammo: A large-scale benchmark dataset for computer-aided diagnosis in
  full-field digital mammography. medRxiv  (2022).
  \doi{10.1101/2022.03.07.22272009}

\bibitem{oh2020deep}
Oh, Y., Park, S., Ye, J.C.: Deep learning covid-19 features on cxr using
  limited training data sets. IEEE transactions on medical imaging
  \textbf{39}(8),  2688--2700 (2020)

\bibitem{rudin2019stop}
Rudin, C.: Stop explaining black box machine learning models for high stakes
  decisions and use interpretable models instead. Nature machine intelligence
  \textbf{1}(5),  206--215 (2019)

\bibitem{rymarczyk2022interpretable}
Rymarczyk, D., Struski, {\L}., G{\'o}rszczak, M., Lewandowska, K., Tabor, J.,
  Zieli{\'n}ski, B.: Interpretable image classification with differentiable
  prototypes assignment. In: European Conference on Computer Vision. pp.
  351--368. Springer (2022)

\bibitem{rymarczyk2021protopshare}
Rymarczyk, D., Struski, {\L}., Tabor, J., Zieli{\'n}ski, B.: Protopshare:
  Prototypical parts sharing for similarity discovery in interpretable image
  classification. In: Proceedings of the 27th ACM SIGKDD Conference on
  Knowledge Discovery \& Data Mining. pp. 1420--1430 (2021)

\bibitem{sacha2024interpretability}
Sacha, M., Jura, B., Rymarczyk, D., Struski, {\L}., Tabor, J., Zieli{\'n}ski,
  B.: Interpretability benchmark for evaluating spatial misalignment of
  prototypical parts explanations. Proceedings of the AAAI Conference on
  Artificial Intelligence  \textbf{38}(1919),  21563–21573 (Mar 2024).
  \doi{10.1609/aaai.v38i19.30154}

\bibitem{cbisddsm}
Sawyer-Lee, R., Gimenez, F., Hoogi, A., Rubin, D.: Curated breast imaging
  subset of digital database for screening mammography (cbis-ddsm) (version 1)
  [data set] (2016), \url{https://doi.org/10.7937/K9/TCIA.2016.7O02S9CY},
  accessed: 28/04/2022

\bibitem{shen2019deep}
Shen, L., Margolies, L.R., Rothstein, J.H., Fluder, E., McBride, R., Sieh, W.:
  Deep learning to improve breast cancer detection on screening mammography.
  Scientific reports  \textbf{9}(1),  1--12 (2019)

\bibitem{shen2021interpretable}
Shen, Y., Wu, N., Phang, J., Park, J., Liu, K., Tyagi, S., Heacock, L., Kim,
  S.G., Moy, L., Cho, K., et~al.: An interpretable classifier for
  high-resolution breast cancer screening images utilizing weakly supervised
  localization. Medical image analysis  \textbf{68},  101908 (2021)

\bibitem{sickles2013acr}
Sickles, E.A., D’Orsi, C.J., Bassett, L.W., Appleton, C.M., Berg, W.A.,
  Burnside, E.S., et~al.: Acr bi-rads{\textregistered} mammography. ACR
  BI-RADS{\textregistered} atlas, breast imaging reporting and data system
  \textbf{5}, ~2013 (2013)

\bibitem{tan2019efficientnet}
Tan, M., Le, Q.: Efficientnet: Rethinking model scaling for convolutional
  neural networks. In: International conference on machine learning. pp.
  6105--6114. PMLR (2019)

\bibitem{wang2023interpretable}
Wang, C., Chen, Y., Liu, F., Elliott, M., Kwok, C.F., Pe{\~n}a-Solorzano, C.,
  Frazer, H., McCarthy, D.J., Carneiro, G.: An interpretable and accurate
  deep-learning diagnosis framework modelled with fully and semi-supervised
  reciprocal learning. IEEE Transactions on Medical Imaging  (2023)

\bibitem{wang2022knowledge}
Wang, C., Chen, Y., Liu, Y., Tian, Y., Liu, F., McCarthy, D.J., Elliott, M.,
  Frazer, H., Carneiro, G.: Knowledge distillation to ensemble global and
  interpretable prototype-based mammogram classification models. In:
  International Conference on Medical Image Computing and Computer-Assisted
  Intervention. pp. 14--24. Springer (2022)

\bibitem{wang2021interpretable}
Wang, J., Liu, H., Wang, X., Jing, L.: Interpretable image recognition by
  constructing transparent embedding space. In: Proceedings of the IEEE/CVF
  international conference on computer vision. pp. 895--904 (2021)

\bibitem{wu2018expert}
Wu, J., Peck, D., Hsieh, S., Dialani, V., Lehman, C.D., Zhou, B., Syrgkanis,
  V., Mackey, L., Patterson, G.: Expert identification of visual primitives
  used by cnns during mammogram classification. In: Medical Imaging 2018:
  Computer-Aided Diagnosis. vol. 10575, pp. 633--641. SPIE (2018)

\bibitem{xu2023sanity}
Xu-Darme, R., Qu{\'e}not, G., Chihani, Z., Rousset, M.C.: Sanity checks for
  patch visualisation in prototype-based image classification. In: Proceedings
  of the IEEE/CVF Conference on Computer Vision and Pattern Recognition. pp.
  3690--3695 (2023)

\end{thebibliography}

\end{document}